\documentclass[conference]{IEEEtran}
\IEEEoverridecommandlockouts
\usepackage{cite}
\usepackage{amsmath,amssymb,amsfonts}
\usepackage{algorithmic}
\usepackage{graphicx}
\usepackage{textcomp}
\usepackage{xcolor}
\usepackage{booktabs}
\usepackage{float}
\usepackage{colortbl}
\usepackage{hyperref}
\def\BibTeX{{\rm B\kern-.05em{\sc i\kern-.025em b}\kern-.08em
    T\kern-.1667em\lower.7ex\hbox{E}\kern-.125emX}}
\begin{document}

\title{E-ARMOR: Edge case Assessment and Review of Multilingual Optical Character Recognition}

\author{\IEEEauthorblockN{Aryan Gupta}
\IEEEauthorblockA{\textit{Intern} \\
\textit{Sprinklr}\\
Gurgaon, India}

\and
\IEEEauthorblockN{Sanskar Soni}
\IEEEauthorblockA{\textit{AI Team} \\
\textit{Sprinklr}\\
Gurgaon, India}
\and
\IEEEauthorblockN{Nuruddin J.}
\IEEEauthorblockA{\textit{AI Team} \\
\textit{Sprinklr}\\
Gurgaon, India}
\and
\IEEEauthorblockN{Shushant Kumar}
\IEEEauthorblockA{\textit{AI Team} \\
\textit{Sprinklr}\\
Gurgaon, India}
\and
\IEEEauthorblockN{Anupam Purwar\IEEEauthorrefmark{1}}
\IEEEauthorblockA{\textit{AI Team} \\
\textit{Sprinklr}\\
Gurgaon, India}
\and
\IEEEauthorblockN{Ratnesh Jamidar}
\IEEEauthorblockA{\textit{AI Team} \\
\textit{Sprinklr}\\
Gurgaon, India}\thanks{\IEEEauthorrefmark{1}Corresponding Author: Anupam Purwar (e-mail: anupam.aiml@gmail.com, https://anupam-purwar.github.io/page/)}

}

\maketitle

\begin{abstract}
Optical Character Recognition (OCR) in multilingual, noisy, and diverse real-world images remains a significant challenge for optical character recognition systems. With the rise of Large Vision-Language Models (LVLMs), there is growing interest in their ability to generalize and reason beyond fixed OCR pipelines. In this work, we introduce Sprinklr-Edge-OCR, a novel OCR system built specifically optimized for edge deployment in resource-constrained environments. We present a large-scale comparative evaluation of five state-of-the-art LVLMs (InternVL, Qwen, GOT OCR, LLaMA, MiniCPM) and two traditional OCR systems (Sprinklr-Edge-OCR, SuryaOCR) on a proprietary, doubly hand annotated dataset of multilingual (54 languages) images. Our benchmark covers a broad range of metrics including accuracy, semantic consistency, language coverage, computational efficiency (latency, memory, GPU usage), and deployment cost. To better reflect real-world applicability, we also conducted edge case deployment analysis, evaluating model performance on CPU only environments. Among the results, Qwen achieved the highest precision (0.54), while Sprinklr-Edge-OCR delivered the best overall F1 score (0.46) and outperformed others in efficiency, processing images 35× faster (0.17 seconds per image on average) and at less than 0.01× of the cost (0.006 USD per 1,000 images) compared to LVLM. Our findings demonstrate that the most optimal OCR systems for edge deployment are the traditional ones even in the era of LLMs due to their low compute requirements, low latency, and very high affordability.

\end{abstract}


\section{Introduction}

Optical Character Recognition (OCR) is a cornerstone technology for digitizing documents, automating data entry, and extracting information from images containing typed, handwritten, or printed text. It  is the process of converting images containing typed, handwritten, or printed text into actual text data.

Traditional Optical Character Recognition (OCR) systems generally follow a multi stage pipeline encompassing four key steps: image pre-processing, layout analysis, character recognition, and post-processing. In the first stage, the input image is enhanced using techniques such as binarization, noise reduction, deskewing, and normalization to improve text clarity. Next, layout analysis or segmentation identifies textual regions, distinguishing them from non-textual elements like graphics, and further segments them into lines, words, and characters. Character recognition follows, where each segmented character is identified using pattern recognition methods, typically involving feature extraction from the bitmap and classification via trained models. Finally, post-processing refines the output by incorporating linguistic context, including spell checking, error correction through n-gram models, and formatting adjustments; for instance, resolving ambiguities such as “an” versus “ar” based on likely language patterns. While effective in controlled environments, these pipelines are often brittle. Their performance degrades significantly with complex layouts, varied fonts, image distortions, and multilingual text, as each stage is prone to cascading errors.

Large Vision-Language Models (LVLMs) mark a significant departure from traditional approaches to visual information processing by enabling joint understanding of images and text within a unified framework. These multimodal models typically integrate a pre-trained vision encoder, such as a Vision Transformer (ViT)\cite{dosovitskiy2020image}, with a Large Language Model (LLM), like LLaMA. The vision encoder first transforms the input image into a sequence of embeddings that capture its visual features. These embeddings are then projected into the same vector space as the LLM’s text embeddings via a specialized projection layer. Both the projected visual features and a user provided textual prompt are passed into the LLM, which generates a response by attending to both visual and textual inputs. Unlike traditional OCR systems that rely on fragile, multistage pipelines for segmentation and character recognition, LVLMs adopt an end-to-end reasoning approach that mimics human like understanding of visual scenes. This paradigm enables them to interpret text in context, offering several advantages: they eliminate the need for explicit character segmentation, support zero shot generalization across diverse languages and fonts, exhibit robustness against real world noise such as glare or skew, and provide prompt based control over outputs. With the help of prompt engineering, these models can be conditioned to generate OCR outputs for the inputted image. However, the high computational requirements of LVLMs often limit their deployment in resource constrained settings. 

Despite these advances, there remains a critical gap in understanding the practical trade offs between traditional OCR systems and LVLM based approaches, particularly in multilingual, noisy, and real world scenarios. Most existing benchmarks focus on monolingual or clean datasets, and very few studies systematically evaluate both accuracy and deployment efficiency such as latency, memory usage, and cost across a wide range of languages and real world conditions\cite{yang2024ccocr}\cite{fu2025ocrbench}. There is a lack of comprehensive, real-world benchmarking that compares the performance and deployment feasibility of state-of-the-art LVLMs and traditional OCR systems on multilingual, noisy, and diverse image datasets, especially in resource-constrained environments.

Edge devices such as Raspberry Pi boards, smartphones, and embedded systems are increasingly used in real world OCR applications due to their portability and cost effectiveness. However, these devices typically operate under severe resource constraints, offering limited RAM (often between 1GiB to 8GiB), modest storage capacity, and low power CPUs or mobile GPUs. Unlike cloud based infrastructure, they lack the memory bandwidth and parallel compute required to run large scale transformer models in real time. As a result, deploying OCR models on such platforms requires careful optimization for latency, memory footprint, and inference efficiency, often through quantization, pruning, and hardware specific acceleration\cite{han2015deep-compression}\cite{kim2021int8mobilegpu}.

In this work, we showcase Sprinklr-Edge-OCR, a compute optimized OCR system suited for edge deployment and inspired by PaddleOCR framework\cite{du2022ppocrv3}. The proposed system incorporates multiple proprietary enhancements over the standard Paddle Structure pipeline, emphasizing modular design, reduced latency, and minimal memory usage. Our model introduces an optimized detection-recognition architecture paired with TensorRT accelerated inference, making it ideal for real time applications on devices with limited resources.

Building on this foundation, we present a comprehensive evaluation of comparison of both traditional OCR engines and state-of-the-art Large Vision-Language Models (LVLMs). Though comprehensive assessment of LLMs is reported \cite{gautam2024opensourceLLMsEnterpriseRAG} and prior studies which benchmark only standard or cloud-based OCR models, our approach is tailored for low-resource environments, enabling efficient and accurate text extraction on devices with limited computational capacity\cite{piryani2025multiocrqa}. To the best of our knowledge, this is the first comprehensive comparative report that systematically evaluates both state-of-the-art Large Vision-Language Models (LVLMs)—including InternVL, Qwen, GOT OCR, LLaMA, and MiniCPM—and traditional OCR systems (including our proposed Sprinklr-Edge-OCR and SuryaOCR) across a proprietary, doubly hand-annotated dataset spanning 54 languages and diverse, noisy, real-world image conditions\cite{paruchuri2025surya}\cite{touvron2023llama}\cite{yao2024minicpm}\cite{zhu2025internvl3}\cite{wei2024general}. Our benchmarking is uniquely novel in its focus on standardized testing for deployment efficiency, encompassing not only accuracy and semantic consistency but also critical metrics such as latency, memory footprint, GPU/CPU usage, and operational cost. Unlike existing research, which typically overlooks the practicalities of low-resource deployment and the impact of model quantization, our study provides actionable insights for practitioners seeking robust OCR solutions in language-rich and computationally constrained settings\cite{singh2025comparativeanalysis}\cite{nagaonkar2025videovlmocr}.

Our results demonstrate that, despite the impressive capabilities of LVLMs, quantized and optimized traditional OCR systems like Sprinklr-Edge-OCR offer superior performance for edge deployment, achieving lower latency, reduced memory usage, and minimal cost without sacrificing accuracy. This work sets a new standard for OCR benchmarking and deployment, bridging the gap between academic research and real-world application in multilingual, resource-limited environments.

\section{Methodology}

\begin{figure} 
    \centering
    \includegraphics[width=1\linewidth]{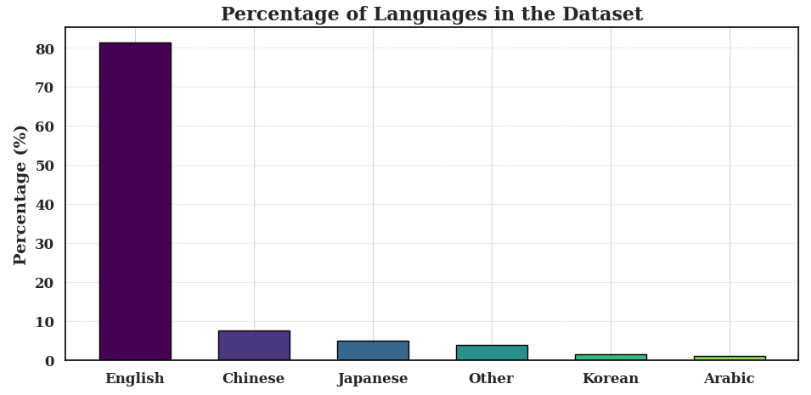}
    \caption{\textbf{Distribution of Languages in the OCR Evaluation Dataset.} The bar chart displays the percentage wise distribution of languages present in the OCR evaluation dataset. English overwhelmingly dominates the dataset, accounting for over 80\% of the total samples. Other languages such as Chinese, Japanese, Korean, and Arabic are represented to a much lesser extent, each comprising less than 10\% of the dataset. The "Other" category includes several low resource and underrepresented languages. This skewed distribution reflects the real world prevalence of English language content in public datasets but also highlights the importance of evaluating model robustness across diverse linguistic contexts, especially in multilingual and non Latin script scenarios.}
    \label{fig:lang-dist}
\end{figure}

\begin{figure} 
    \centering
    \includegraphics[width=1\linewidth]{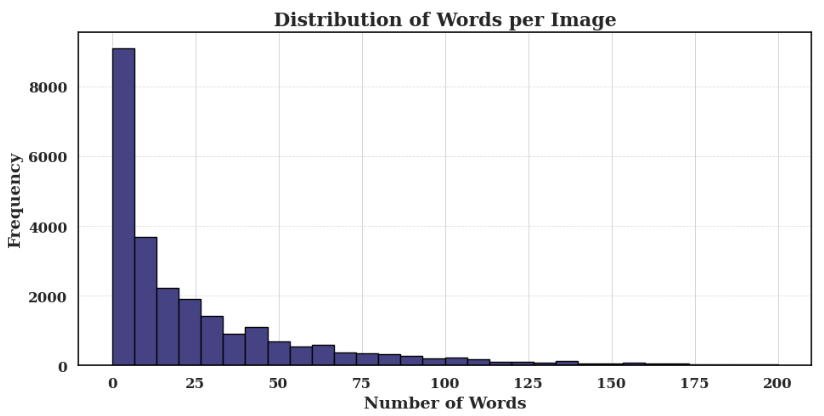}
    \caption{\textbf{Distribution of Word Counts per Image in the Dataset.} The histogram shows the distribution of the number of words per image across the dataset. The majority of images contain a relatively small number of words, with a sharp peak at fewer than 10 words per image, and a long tail extending beyond 150 words.}
    \label{fig:num-words}
\end{figure}

We curated a challenging, real-world dataset comprising of images with 54 distinct languages, with the majority of the images containing text written in English, Chinese, Japanese, Korean and Arabic (see Figure~\ref{fig:lang-dist}). The dataset has a high diversity of content, which includes posters, city view, memes, screenshots, and advertisements containing multilingual text, complex layouts, and various visual artifacts. To establish a high quality ground truth for evaluation, each image was doubly annotated by humans. Figure~\ref{fig:num-words} illustrates the distribution of word counts per image, most images contain fewer than 10 words, with a long tail extending past 150. The distribution is highly skewed, indicating that while most images are text sparse (e.g., signs, labels, short messages), a smaller subset includes dense textual content such as documents, tables, or webpages. This variability poses challenges for OCR models, which must generalize across both minimal and high-density text scenarios.

\subsection{Models}
\begin{table}[h!]
\centering
\caption{\textbf{Overview of Popular Open Source OCR and LVLMs.} The table summarizes a range of widely used open source OCR models, indicating whether they supports more than four languages.}
\begin{tabular}{|l|c|}
\hline
\textbf{Name} & \textbf{Supports more than 4 Languages?} \\
\hline
TesseractOCR (CPU only)\cite{smith2007overview} & Yes \\
MMOCR\cite{kuang2021mmocr} & No \\
TrOCR\cite{li2021trocr} & No \\
DOCTR\cite{liao2023doctr} & No \\
PARSEQ\cite{bautista2022scene} & No\\
ABINet\cite{fang2021read} & No  \\
Smoldocling\cite{nassar2025smoldocling} & No \\
Bridging-Text-Spotting\cite{huang2024bridginggapendtoendtwostep} & No\\
PyLaila\cite{pylaila} & No\\
Kraken\cite{kiessling2019kraken} (historical languages) & No \\
InstructOCR\cite{instructocr} & No\\
Paddle Structure\cite{du2022ppocrv3} & Yes\\
Sprinklr-Edge-OCR & Yes\\
Surya\cite{paruchuri2025surya} & Yes\\
QWEN VL\cite{bai2023qwen} & Yes \\
Llama 3.2 V\cite{touvron2023llama} & Yes \\
GOT OCR 2.0\cite{wei2024general} & Yes \\
InternVL\cite{zhu2025internvl3} & Yes \\
MiniCPM-V-2.6\cite{yao2024minicpm} & Yes \\
\hline
\end{tabular}
\end{table}

The models listed in Table 1 were selected to represent a diverse cross section of state-of-the-art open source OCR systems, spanning both traditional and modern deep learning approaches. Selection criteria included multilingual support, architectural diversity (from lightweight pipelines to large multimodal models), availability of pretrained weights. This variety ensures a comprehensive evaluation across different document types, languages, and deployment scenarios.



\begin{figure*}[htb]
    \centering
    \includegraphics[width=1\linewidth]{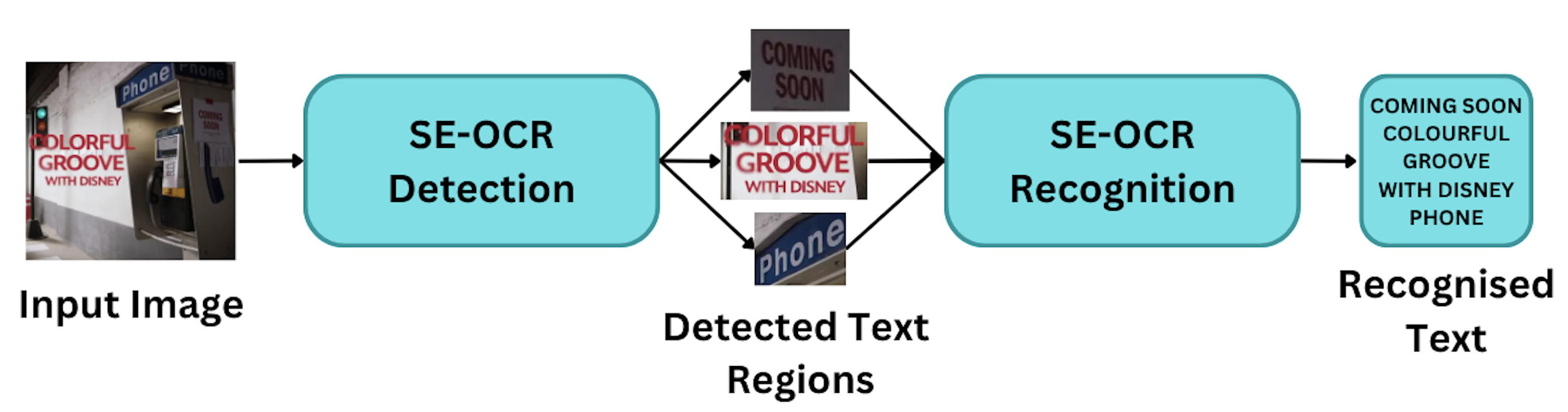}
    \caption{\textbf{SE-OCR system pipeline for text extraction.} The SE-OCR framework operates in two stages: text detection and text recognition. Given an input image containing text, the detection module first localizes individual text regions. These cropped regions are then passed to the recognition module, which transcribes the content into machine readable text. The final output is a sequence of recognized text tokens, accurately reflecting the textual information present in the original image.}
    \label{fig:SE-OCR}
\end{figure*}

Sprinklr-Edge-OCR engine is built upon the PaddleOCR framework, specifically optimized for edge deployment\cite{du2022ppocrv3}. Unlike the traditional Paddle-Structure-v3 pipeline, which employs a comprehensive suite of modules, including layout detection, chart and table engines, and document orientation classification, our approach focuses on essential OCR tasks and integrates several proprietary modifications aimed at optimizing performance for resource constrained environments. These proprietary enhancements are based on recent advancements in lightweight and modular OCR design, resulting in a compact and efficient system (see Figure~\ref{fig:SE-OCR}). To further boost run-time performance, the engine integrates TensorRT for accelerated inference on supported hardware. The engine supports multiple languages, including simplified Chinese, Chinese Pinyin, Traditional Chinese, English, and Japanese, with extensibility to others through fine tuning. Designed for high-throughput, latency-sensitive edge applications. Sprinklr-Edge-OCR delivers strong recognition accuracy while maintaining a minimal computational footprint and delivers a highly compact, efficient, and scalable solution. The result is a state-of-the-art OCR system that is ideally suited for high-throughput, latency-sensitive applications at the edge, without sacrificing recognition accuracy.



The datalab's toolkit SuryaOCR is a comprehensive, open‑source document analysis suite that offers OCR, layout analysis, reading order detection, and table recognition in more than 90 languages. It leverages a SegFormer based detection model and a DONUT based recognition backbone\cite{xie2021segformer}\cite{kim2021ocr}, delivering comparable performance to traditional engines like Tesseract and cloud services. It supports line-level text detection, structural region classification (e.g., headers, images, tables), and multi-column, right-to-left reading flow reconstruction. Outputs include richly annotated JSON with bounding boxes, polygons, confidence scores, and semantic region labels, as well as optional visual overlays.


The General OCR Theory 2.0 (GOT OCR 2.0) is a unified, end-to-end model designed to handle a diverse range of artificial optical signals, including plain text, mathematical formulas, molecular structures, tables, charts, sheet music, and geometric shapes. Developed by Haoran Wei et al., GOT OCR 2.0 uses a high compression encoder and a long-context decoder to process and generate output from visual input. The encoder compresses 1024×1024 pixel images into 256 tokens of size 1024, while the decoder, based on the Qwen-0.5B model, can handle long contexts up to 8,000 tokens. Key features of GOT OCR 2.0 include interactive OCR with region-level recognition guided by coordinates or colors, dynamic resolution handling for ultra-high-resolution images, multi-page OCR capabilities, and support for various output formats such as plain text, Markdown, TikZ, SMILES notation, and LaTeX. These advancements make GOT OCR 2.0 a versatile tool for complex OCR tasks across different domains.


The OpenGVLab-InternVL2.5-1B is a multimodal foundation model designed for a wide range of vision-language tasks. It employs a "ViT-MLP-LLM" architecture, integrating a Vision Transformer (InternViT-300M-448px-V2\_5) for image processing, a Multilayer Perceptron (MLP) projector for cross-modal alignment, and a Large Language Model (Qwen2.5-0.5B-Instruct) for text understanding. This design allows the model to process and understand single images, multi-image datasets, and videos, with inputs resized to 448x448 pixels and normalized using ImageNet statistics \cite{deng2009imagenet}. InternVL2.5-1B has demonstrated strong performance across various benchmarks, including OCR, chart, and document understanding, as well as visual grounding and video comprehension tasks. Its architecture enables efficient processing and understanding of complex multimodal data, making it a versatile tool for applications requiring integrated vision and language capabilities.


The Unsloth/Llama-3.2-11B-Vision-Instruct 4Bit is a vision language model developed by Meta and optimized by Unsloth. It combines an 11-billion-parameter Llama 3.2 architecture with a vision adapter, enabling the model to process and understand both visual and textual inputs. The model utilizes Unsloth's Dynamic 4-bit quantization, achieving approximately 2x faster inference and 60\% reduced memory usage compared to the original implementation. This optimization allows for efficient deployment on consumer grade GPUs. The architecture includes a Vision Transformer (ViT) for image processing, cross-attention layers to integrate visual and textual information, and a language model to generate responses. It supports a wide range of vision-language tasks, including image captioning, visual question answering, and multimodal reasoning. In addition, it supports long context lengths, which improves its ability to handle extended conversations and complex visual inputs.


The openbmb/MiniCPM-V-2\_6-int4 is a compact 8 billion-parameter multimodal large language model (MLLM) optimized for edge devices. It integrates components from SigLip-400M, and Qwen2-7B, delivering robust performance across vision, and language tasks\cite{zhai2023sigmoid}. We employ 4-bit quantization to enhance efficiency, enabling deployment on consumer grade hardware. It supports high resolution image processing and optical character recognition (OCR). Additionally, MiniCPM-V-2\_6-int4 offers real time continuous video/audio processing, multilingual support for English, Chinese, German, French, Italian, Korean, etc. Its versatility and efficiency, extremely low token density (i.e., number of pixels encoded into each visual token) make it suitable for a wide range of applications, including real time conversation, and multimodal streaming.




The JackChew/Qwen2-VL-2B-OCR is a specialized multimodal large language model (MLLM) developed to enhance optical character recognition (OCR) capabilities. It is a fine tuned version of Qwen2-VL-2B-Instruct, which employs a Vision Transformer (ViT - 600 M params) for image processing and QWEN2 large language models for text understanding. The model utilizes a dynamic resolution mechanism called Naive Dynamic Resolution support, allowing it to process images of varying sizes and complexities, thereby improving recognition accuracy across diverse image types and sizes. Fine tuning ensures \texttt{Qwen2-VL-2B-OCR} particularly is more effective for applications requiring detailed document analysis, including form processing, invoice extraction, and multilingual text recognition.

\subsection{Experimental Setup}
The following LVLMs and OCR models were evaluated: InternVL, Qwen, GOT, LLaMA, MiniCPM, Sprinklr-Edge-OCR and SuryaOCR. Each model was provided with the same images, and the models that require prompt inputs had the same standard prompt.

\begin{figure}[H]
    \centering
    \includegraphics[width=1\linewidth]{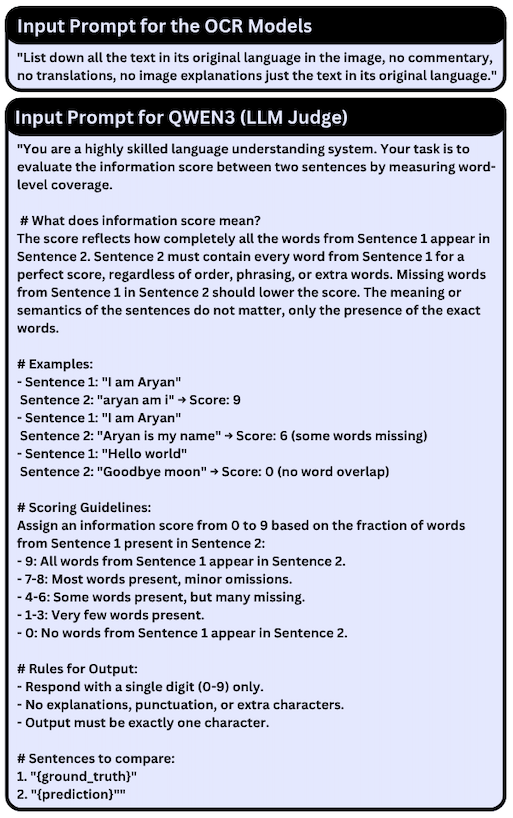}
\end{figure}

All of the above mentioned models were benchmarked on NVIDIA T4 GPUs, along with 4 vCPUs (Intel Xeon Family (upto 2.5 GHz)) and 16 GiB of RAM. 

\begin{figure*}[htb]
    \centering
    \includegraphics[width=1\linewidth]{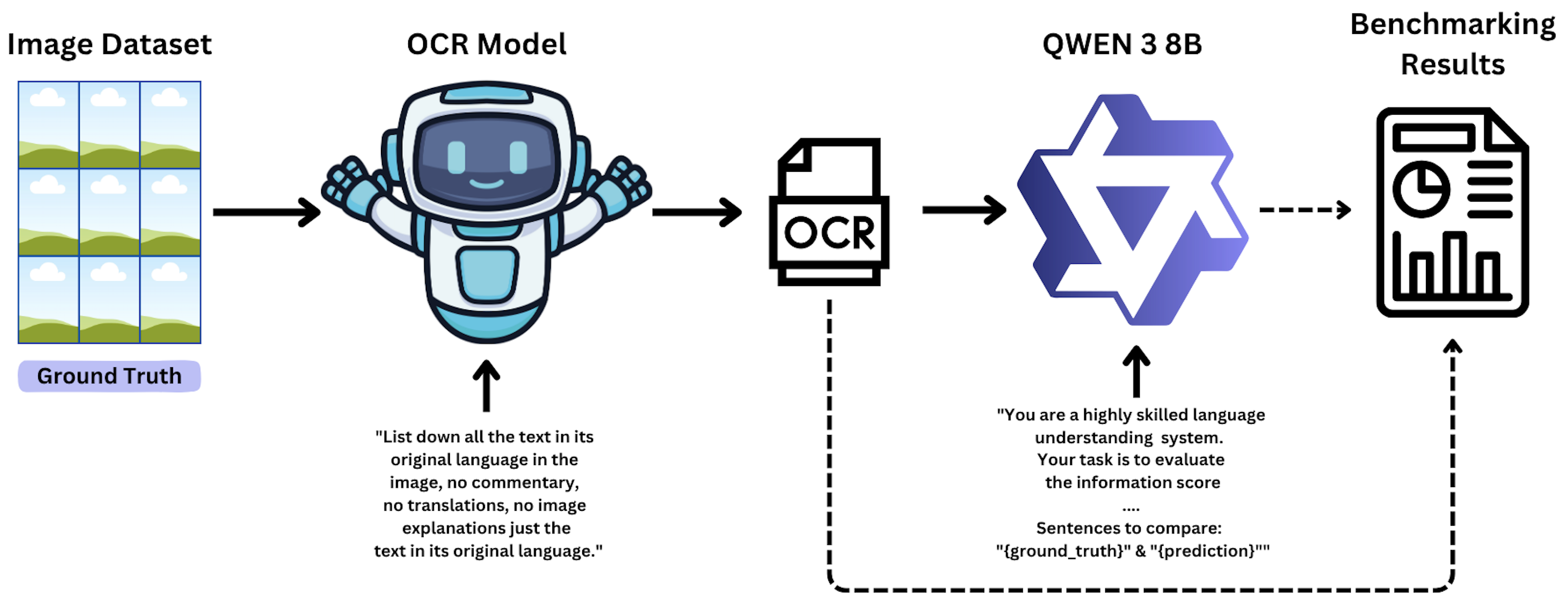}
    \caption{\textbf{End-to-end benchmarking pipeline for OCR model evaluation.} The pipeline begins with an image dataset annotated with ground truth text. The OCR model is prompted to extract the text from each image in its original language. These OCR predictions are then evaluated using the Qwen 3 8B language model, which is prompted as a language understanding system to compare the predicted text against the ground truth. The model assigns an information similarity score. Raw outputs and ground truth combined with the similarity score is then used to generate benchmarking metrics. }
    \label{fig:pipeline}
\end{figure*}

The process begins with the multilingual image dataset paired with human annotated ground truth annotations. These images are passed through an OCR model, which is prompted to extract all visible text exactly as it appears. The OCR model’s output is then assessed by Qwen 3 (8B), a large language model, which is instructed to compare the OCR prediction with the ground truth and assign a similarity accuracy score. This comparison helps quantify similarity and extraction fidelity. The final benchmarking results are compiled to evaluate model performance comprehensively (see Figure~\ref{fig:pipeline}).

\subsection{Evaluation Metrics}
To ensure a comprehensive evaluation, we employed a wide range of metrics, each providing a unique perspective on model performance. The Similarity Score (LLM Judge) metric employs the Qwen3-8B language model to evaluate word level similarity between the ground truth and the predicted text\cite{qwenteam2025qwen3}\cite{zheng2023judging}, as LLM as judge have shown great efficacy in evaluating ML model outputs \cite{harbola2025knowslm}. A handcrafted prompt is provided to the model, instructing it to ignore word order, semantics, grammar, and extra content, and to focus solely on the presence of exact ground truth words within the prediction. The model returns a single digit score from 0 to 9, where:
\begin{itemize}
    \item \textbf{9} indicates that all ground truth words are present in the prediction.
    \item \textbf{7--8} indicates most words are present, with minor omissions.
    \item \textbf{4--6} indicates partial word overlap.
    \item \textbf{1--3} indicates only a few ground truth words are retained.
    \item \textbf{0} indicates no overlap at all.
\end{itemize}
This evaluation strategy is designed to capture nuances that traditional metrics often miss, such as spelling variations, translation errors, and word order. It reflects how much of the original textual content is in the predicted output, making it particularly useful for evaluating multilingual OCR models. Higher scores indicate stronger alignment with the source words, while lower scores highlight significant omissions or hallucinations.

\section{Results}

The Sprinklr-Edge-OCR pipeline achieved a substantial reduction in latency, from 0.57 seconds to 0.17 seconds, and peak VRAM usage decreased from 9.7 GiB to 1.8 GiB when compared to the traditional Paddle structure pipeline. To validate competitive performance of the novel model, we ran benchmarking on sub-datasets from OCRBench v2\cite{fu2025ocrbench}. We evaluated our model on OCRBench v2's English text recognition tasks, which included 3 datasets: full-page OCR, fine grained text recognition, and general text recognition. Across these three benchmarks, our model achieved an average score of 55.4/100, demonstrating competitive performance. This places it on par with many highly regarded multimodal models such as GPT-4o-mini and InternVL2.5-8B as seen in the OCRBench v2 leaderboard. Despite being lightweight and efficient, the model matches or outperforms several significantly larger models with more complex architectures.

The results highlight a contrast in edge deployment efficiency. Sprinklr-Edge-OCR outperformed Qwen-VL by more than 15× in terms of inference speed and used nearly 12× less memory. While Qwen-VL's performance is constrained by its large model size and dependency on extensive compute resources, Sprinklr-Edge-OCR's streamlined architecture demonstrates clear advantages for low latency, low footprint deployments on CPU.

These findings reinforce Sprinklr-Edge-OCR’s suitability for real time applications on edge devices such as embedded systems, kiosks, or offline document scanners contexts where latency, memory constraints, and power efficiency are critical.

For example, as demonstrated in our experiments, prompts can effectively guide the model to extract only the original language text, thereby avoiding unnecessary translations or image descriptions. The performance of each LVLM across the evaluated metrics is summarized in Table 1. The best-performing value for each metric is highlighted in bold. The cost calculations were based on the G4dn.xlarge AWS instance, which includes a T4 Tensor Core GPU, 4 vCPUs and 16 GiB of RAM. This instance is currently priced at \$0.526 per hour under the On-Demand pricing model.

\begin{table*}[htb]
\centering
\caption{\textbf{Evaluation of OCR and Vision-Language Models Across Multiple Dimensions.} The table summarizes the performance of various models across three categories: Error Metrics, Accuracy Metrics, and Resource Consumption. This comprehensive comparison enables analysis of trade offs between recognition quality and deployment efficiency.}
\setlength{\tabcolsep}{10pt} 
\begin{tabular}{@{}lccccccc@{}}
\toprule
\textbf{Metric} & \textbf{InternVL} & \textbf{Qwen} & \textbf{GOT} & \textbf{LLaMA} & \textbf{MiniCPM} & \textbf{Sprinklr-Edge-OCR} & \textbf{Surya} \\ \midrule
\rowcolor[gray]{0.9} \multicolumn{8}{c}{\textbf{Error Metrics (Lower is Better)}} \\
WER & 4.8320  & 0.8609 & 0.9875 & 2.7893 & 2.9658 & \textbf{0.8528} & 1.2482 \\
CER & 6.7943  & 1.0802 & \textbf{0.6459} & 3.9395 & 4.6542 & 0.6713 & 2.9562 \\
Avg Levenshtein Dist & 268.58  & 122.52 & 138.69 & 184.54 & 245.92 & \textbf{109.94} & 194.39 \\
Avg Missed Words & 26.12  & 23.10 & 22.10 & 22.10 & 19.49 & \textbf{18.48} & 24.09 \\
Avg Extra Words & 29.09  & \textbf{4.03} & 11.85 & 17.98 & 21.69 & 11.95 & 11.48 \\
Mean Per-Word Levenshtein & 2.6038  & 6.0008 & 2.2581 & 1.7896 & 1.7423 & \textbf{1.4949} & 8.0993 \\
\midrule
\rowcolor[gray]{0.9} \multicolumn{8}{c}{\textbf{Accuracy Metrics (Higher is Better)}} \\
F1 Score & 0.1545  & 0.3690 & 0.3675 & 0.3193 & 0.3804 & \textbf{0.4570} & 0.2357 \\
Precision & 0.1573  & \textbf{0.5426} & 0.4511 & 0.3445 & 0.3833 & 0.5010 & 0.3013 \\
Recall & 0.1808  & 0.3149 & 0.3493 & 0.3467 & 0.4134 & \textbf{0.4398} & 0.2058 \\
Similarity & 4.8  & 6.0 & 5.8 & 5.7 & 6.6 & \textbf{7.2} & 6.5 \\
\midrule
\rowcolor[gray]{0.9} \multicolumn{8}{c}{\textbf{Resource Consumption (Lower is Better)}} \\
Params (B) & 1  & 2 & 0.58 & 11 & 8 & \textbf{0.15} & 0.52 \\
Avg Time (s) & 7.02  & 5.83 & 3.64 & 10.21 & 13.21 & \textbf{0.17} & 1.16 \\
Max Time (s) & 28.67  & 31.59 & 151.07 & 62.07 & 258.61 & \textbf{6.32} & 32.62 \\
Max Memory (MiB) & 14239  & 12907 & 5857 & 8709 & 9761 & \textbf{1970} & 7471 \\
Avg GPU Util (\%) & 42.57  & 89.76 & 31.16 & 78.41 & \textbf{91.26} & 2.07 & 3.51 \\
Number of Threads & 1 & 1 & 2 & 1 & 1 & 4 & 2 \\
Cost per 1000 images (\$) & 1.02 & 0.85 & 0.27 & 1.49 & 1.93 & \textbf{0.006} & 0.08 \\
\bottomrule
\end{tabular}
\end{table*}

\begin{figure*}[htb]
    \centering
    \includegraphics[width=1\linewidth]{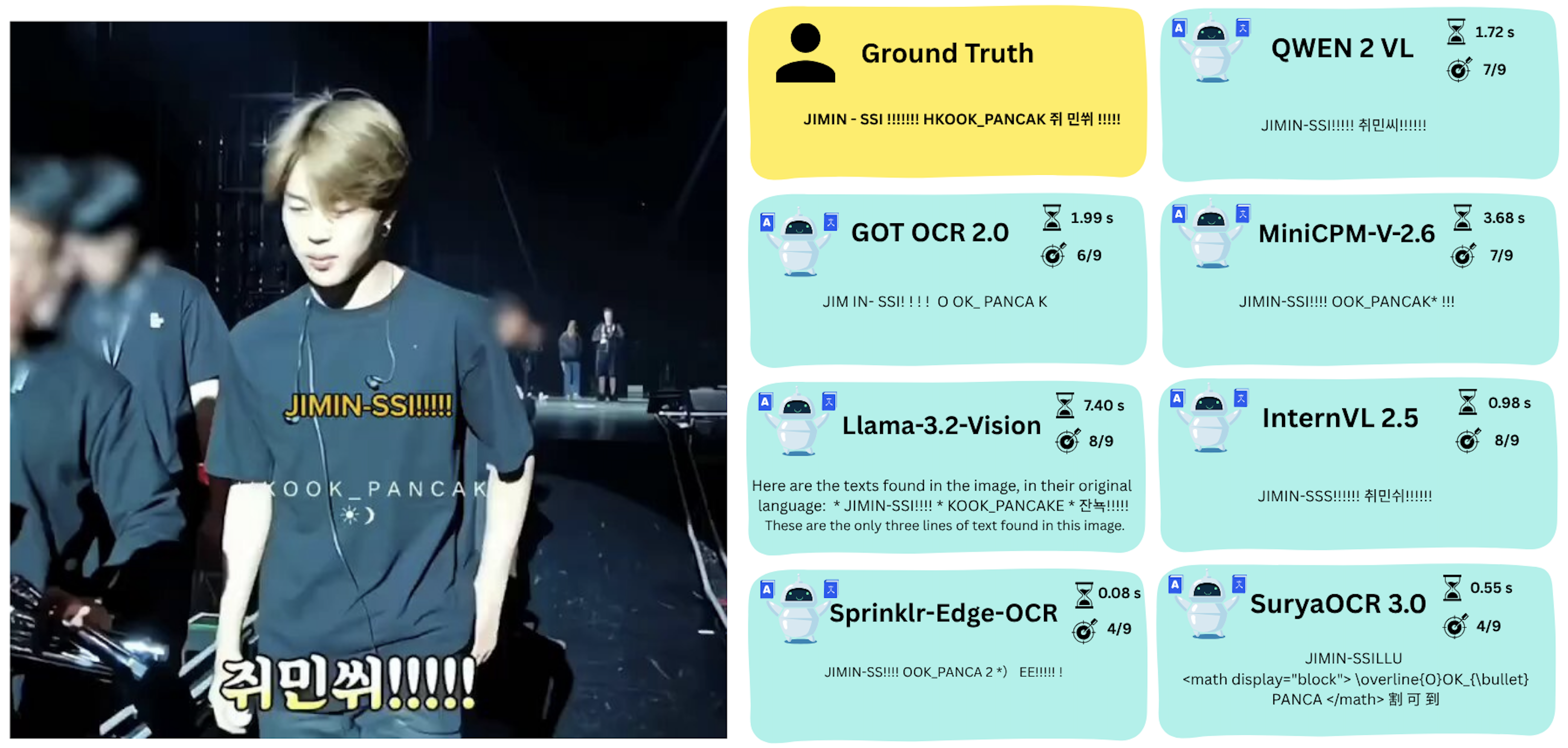}
    \caption{Example outputs for a publicly available challenging image, time taken and similarity score of various models along with ground truth}
    \label{fig:examples}
\end{figure*}

Figure~\ref{fig:examples} presents representative output examples generated by each of the evaluated models. Table 2 presents a comprehensive comparison of seven OCR systems across three key dimensions: error metrics, accuracy, and resource consumption. Among traditional OCR systems. Sprinklr-Edge-OCR consistently outperformed others, achieving the lowest Word Error Rate (WER = 0.8528) and second-best Character Error Rate (CER = 0.6713), while also recording the lowest average Levenshtein distance (109.94) and fewest missed words (18.48). Qwen VL, a Large Vision-Language Model (LVLM), showed high precision (0.5426) and the lowest average extra words (4.03), indicating strong text extraction discipline.

\begin{figure}
    \centering
    \includegraphics[width=1\linewidth]{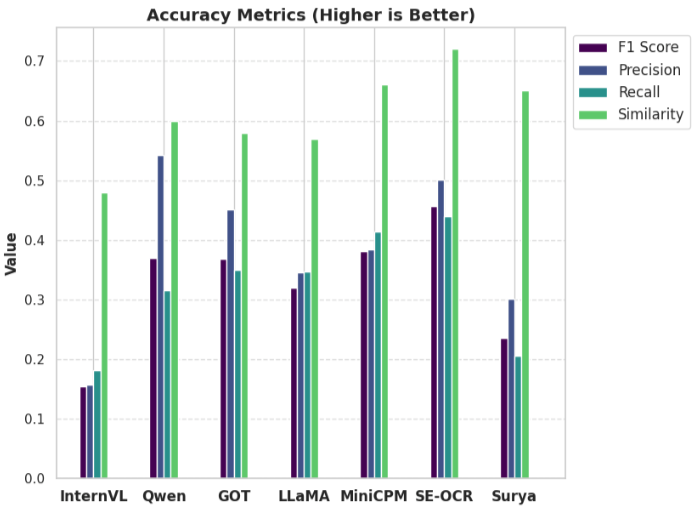}
    \caption{\textbf{Comparison of Accuracy Metrics Across OCR Models.} This bar plot presents the performance of the seven selected OCR and vision-language models across four key accuracy metrics: F1 Score, Precision, Recall, and Similarity score. These metrics reflect the models’ ability to accurately recognize and reproduce textual content from images. Higher values indicate better performance. Among the models, SE-OCR consistently ranks highest across most metrics, notably achieving a Similarity score exceeding 7, indicating high semantic closeness between predicted and ground truth text. QWEN showcases the strongest performance in terms of precision. MiniCPM and GOT also perform well, with balanced precision and recall. In contrast, InternVL shows the lowest performance across all accuracy metrics, highlighting its limitations in text specific recognition tasks.}
    \label{fig:accuracy}
\end{figure}

\begin{figure}
    \centering
    \includegraphics[width=1\linewidth]{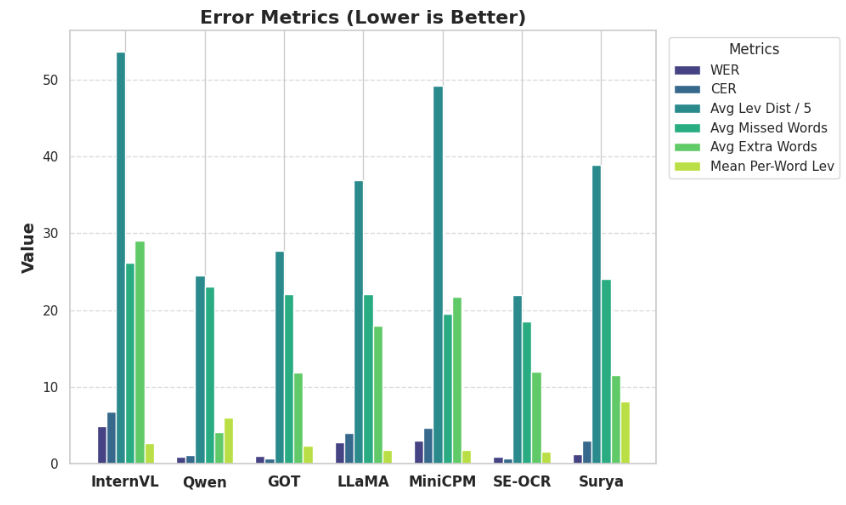}
    \caption{\textbf{Comparison of Error Metrics Across OCR Models (Lower is Better).} The figure illustrates six key error metrics: Word Error Rate (WER), Character Error Rate (CER), Average Levenshtein Distance (scaled), Average Missed Words, Average Extra Words, and Mean Per-Word Levenshtein Distance for the seven selected OCR and vision-language models. These metrics quantify different aspects of recognition failure, including character level and word level discrepancies between predicted and ground truth text. SE-OCR consistently achieves the lowest error values across most metrics, reflecting high robustness and recognition precision. QWEn also showcases an extremely low error rate. GOT and MiniCPM also exhibit relatively low error rates, with balanced performance on both missed and extra words. In contrast, InternVL and Surya show significantly higher error rates across all dimensions, indicating poor alignment with ground truth text and increased frequency of both omissions and insertions.}
    \label{fig:errors}
\end{figure}

For overall accuracy, Sprinklr-Edge-OCR achieved the highest F1 Score (0.4570), demonstrating balanced performance in both precision and recall. MiniCPM achieved the highest recall (0.4134), though at the cost of extra words and slightly reduced precision. GOT OCR showed the best CER (0.6459), revealing strength in character-level recognition (see Figure~\ref{fig:errors} \& ~\ref{fig:accuracy}).

In terms of semantic similarity, Sprinklr-Edge-OCR again led with a similarity score of 7.2, outperforming all other models.

\begin{figure}
    \centering
    \includegraphics[width=1\linewidth]{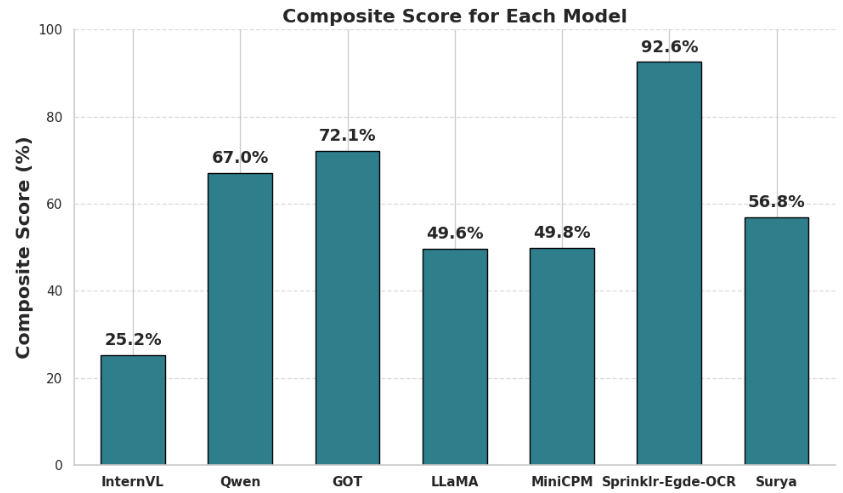}
    \caption{\textbf{Composite Score Comparison Across OCR Models.} The bar chart illustrates the normalized composite scores (out of 100\%) for the seven OCR and vision-language models. Among all models, Sprinklr-Edge-OCR achieves the highest composite score at 92.6\%, indicating strong overall performance in both accuracy and deployment efficiency. In contrast, InternVL ranks lowest at 25.2\%, reflecting limitations in real world robustness and efficiency. This visualization also highlights the substantial performance variability across modern OCR systems.}
    \label{fig:comp}
\end{figure}

The composite score was calculated by aggregating multiple evaluation metrics, including error rates (e.g., WER, CER), accuracy measures (e.g., F1 score, precision, recall), and similarity into a single normalized score ranging from 0 to 1, later scaled to 100 for readability. Each metric was first normalized across all models, with higher is better metrics scaled directly and lower is better metrics inverted. The final score represents an average of these normalized values, providing a balanced view of both recognition quality and robustness. As shown in the results, Sprinklr-Edge-OCR outperformed all other models with a score of 92.6, followed by GOT (72.1) and Qwen (67.0). Models like InternVL scored significantly lower, highlighting trade offs between model type and OCR performance in real-world conditions (see Figure~\ref{fig:comp}).

\begin{figure}
    \centering
    \includegraphics[width=1\linewidth]{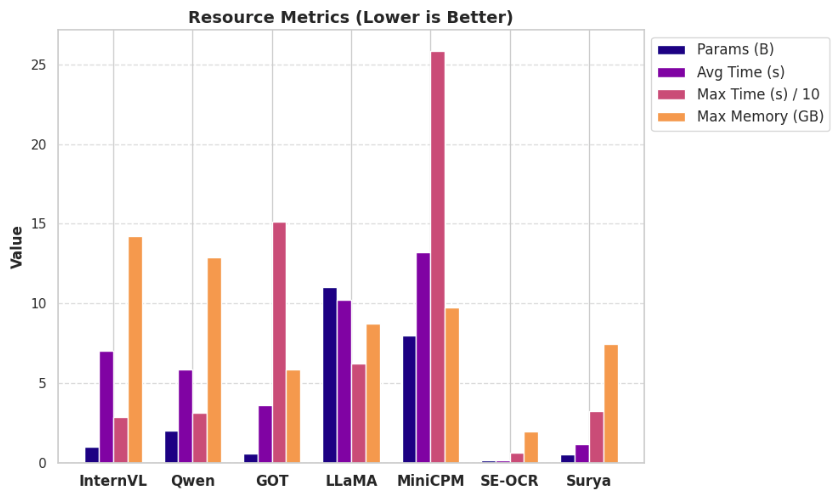}
    \caption{\textbf{Comparison of Resource and Efficiency Metrics Across OCR Models (Lower is Better).} The chart presents the computational efficiency of the seven selected OCR and vision language models, evaluated using four key resource related metrics: Model Size (in billions of parameters), Average Inference Time (in seconds), Maximum Inference Time (scaled by 1/10), and Peak Memory Usage (in GiB). Lower values across these metrics indicate better deployment efficiency. SE-OCR demonstrates exceptional efficiency, requiring minimal memory and computation time, and using significantly fewer parameters than LLM based counterparts. In contrast, MiniCPM shows the highest resource consumption, particularly in maximum inference time (peaking above 250 seconds before scaling) and memory usage. InternVL, Qwen, and LLaMA also incur high memory and parameter costs.}
    \label{fig:efficiency}
\end{figure}

Sprinklr-Edge-OCR is also notable for  the fastest average inference time (0.17 seconds), lowest maximum memory usage (1970 MiB), and lowest cost per 1,000 images (\$0.006), making it highly suitable for real-time edge deployment. In contrast, LVLMs like MiniCPM and LLaMA consumed significantly more time, memory, and compute resources. For instance, MiniCPM's average inference time was 13.21 seconds and peak memory usage exceeded 9.7 GiB (see Figure~\ref{fig:efficiency}).

SuryaOCR offered moderate performance, better than InternVL and LLaMA in error metrics, but lagged behind Sprinklr-Edge-OCR in both speed and accuracy.

\subsection{Edge Deployment Analysis on CPU}

To assess the feasibility of real world deployment in resource constrained environments, we conducted CPU only inference benchmarking on a system equipped with an 8 core Intel Xeon 8375C (Ice Lake, 3.5 GHz) processor and 64 GiB of RAM. This setting simulates an edge deployment scenario where GPU access is unavailable.

We evaluated two contrasting, top-performing models: the large vision-language model QwenVL, and the lightweight, updated OCR pipeline Sprinklr-Edge-OCR (focused solely on detection and recognition). Table 3 summarizes average inference time per image and peak RAM usage during runtime


\begin{table}[htb]
\centering
\scriptsize
\caption{CPU-Only Edge Deployment Performance Comparison}
\begin{tabular}{@{}lcc@{}}
\toprule
\textbf{Metric}              & \textbf{Sprinklr-Edge-OCR} & \textbf{Qwen-VL} \\ \midrule
Avg Inference Time (s)       & \textbf{4.36}               & 69.38           \\
Time Cost (×)                & 1.0                         & 15.9×          \\
Peak RAM Usage (GiB)         & \textbf{0.89}               & 10.8           \\
Ram Required (×)                & 1.0                         & 12.1×          \\
\bottomrule
\end{tabular}
\end{table}

\section{Discussion}

Our analysis reveals key qualitative insights into the behavior and trade-offs of Large Vision-Language Models (LVLMs) versus traditional OCR pipelines. LVLMs such as Qwen offer distinct advantages in multilingual and zero-shot settings, where broader contextual reasoning and semantic alignment are essential. For instance, Qwen consistently produced output with high textual fidelity and minimal hallucinations, making it particularly suitable for applications where precision and strict adherence to source content are critical without any restrictions on available compute.

Hybrid and emerging models like GOT OCR and MiniCPM showcased notable individual strengths, such as GOT’s lowest Character Error Rate (CER), but struggled to deliver consistent performance across all evaluation metrics. This suggests that while LVLMs and hybrid models can excel in specialized or multitask scenarios, they may not yet match the end to end reliability of established OCR systems.

Sprinklr-Edge-OCR, despite being based out of traditional Machine learning pipeline supports 5 languages, remained highly competitive. It outperformed or matched LVLMs on most core metrics, including F1 score, latency, and memory usage, making it particularly well suited for real world, resource constrained deployments. This highlights the continued relevance and strength of optimized, task specific OCR pipelines, especially when reliability and deployment efficiency are paramount.

To further evaluate the practicality of deploying these models in edge environments, we conducted CPU-only inference tests simulating scenarios without GPU acceleration. Using an 8-core Intel Xeon processor, we compared the large vision-language model Qwen-VL with the lightweight Sprinklr-Edge-OCR pipeline. The results demonstrate a stark contrast in resource requirements: Qwen-VL incurred significantly higher inference latency (69.38 seconds per image) and memory usage (10.8 GiB RAM), reflecting the computational demands of LVLMs. In contrast, Sprinklr-Edge-OCR achieved rapid inference (4.36 seconds per image) with minimal memory consumption (0.89 GiB RAM), underscoring its suitability for edge deployment. These findings reinforce that while LVLMs offer advanced capabilities, traditional OCR pipelines like Sprinklr-Edge-OCR remain far more efficient and practical for real-world applications on resource-constrained hardware.

\section{Conclusion}

This study provides a comprehensive evaluation of multilingual OCR models across a range of scenarios, including edge deployment and real world image complexity. The key finding is that there is no one size fits all solution. The optimal model depends on the intended application and deployment context.

LVLMs like Qwen offer compelling strengths in semantic reasoning, language generalization, and zero shot adaptability. While they currently lag behind in metrics such as latency and overall F1 score, they hold promise for next generation OCR systems that demand deeper contextual understanding. However, their current computational demands make them unsuitable for deployment on edge devices.

Conversely, for applications where efficiency, scalability, and low latency are critical, such as on device or edge environments, Sprinklr-Edge-OCR emerges as the top choice as it consistently delivering the best overall accuracy and performance with minimal computational overhead.

In sum, this work lays a practical evaluation of OCR models, balancing accuracy, cost, and adaptability, and guiding the development of systems for increasingly complex multilingual and real world OCR scenarios.

\end{document}